%% file: main.tex
\documentclass[10pt,twocolumn,letterpaper]{article}

\usepackage{cvpr}              


\definecolor{cvprblue}{rgb}{0.21,0.49,0.74}
\usepackage[pagebackref,breaklinks,colorlinks,allcolors=cvprblue]{hyperref}

\usepackage[ruled]{algorithm2e} 
\usepackage{cuted}
\usepackage{times}
\usepackage{makecell} 
\usepackage{graphicx} 
\usepackage{amsmath}  
\usepackage{multirow} 

\newcommand{\methodname}{{DynamicScaler}\xspace}

\def\paperID{6593} 
\def\confName{CVPR}
\def\confYear{2025}

\title{\methodname: Seamless and Scalable Video Generation for Panoramic Scenes}

\author{
    Jinxiu Liu\textsuperscript{1,}\thanks{Equal contribution.} \;
    Shaoheng Lin\textsuperscript{1,*} \;
    Yinxiao Li\textsuperscript{2,}\thanks{Equal advising} \;
    Ming-Hsuan Yang\textsuperscript{2,3,\dag}
    \\
    \textsuperscript{1}South China University of Technology \quad
    \textsuperscript{2}Google DeepMind \quad
    \textsuperscript{3}UC Merced
}

\begin{document}
\maketitle

\input{sec/0_abs}

\input{sec/1_intro}

\input{sec/2_relatedwork}

\input{sec/3_method}
\input{sec/4_exp}
{
    \small
    \bibliographystyle{ieeenat_fullname}
    \bibliography{main}
}

\end{document}

%% file: sec/0_abs.tex
\begin{abstract}

The increasing demand for immersive AR/VR applications and spatial intelligence has heightened the need to generate high-quality scene-level and 360° panoramic video. However, most video diffusion models are constrained by limited resolution and aspect ratio, which restricts their applicability to scene-level dynamic content synthesis. In this work, we propose \textbf{DynamicScaler}, addressing these challenges by enabling spatially scalable and panoramic dynamic scene synthesis that preserves coherence across panoramic scenes of arbitrary size. Specifically, we introduce a Offset Shifting Denoiser, facilitating efficient, synchronous, and coherent denoising panoramic dynamic scenes via a diffusion model with fixed resolution through a seamless rotating Window, which ensures seamless boundary transitions and consistency across the entire panoramic space, accommodating varying resolutions and aspect ratios. Additionally, we employ a Global Motion Guidance mechanism to ensure both local detail fidelity and global motion continuity. Extensive experiments demonstrate our method achieves superior content and motion quality in panoramic scene-level video generation, offering a training-free, efficient, and scalable solution for immersive dynamic scene creation with constant VRAM consumption regardless of the output video resolution. Project page is available at \url{https://dynamic-scaler.pages.dev/new} .

\begin{figure}[!h]
    \centering
    \includegraphics[width=0.5\textwidth]{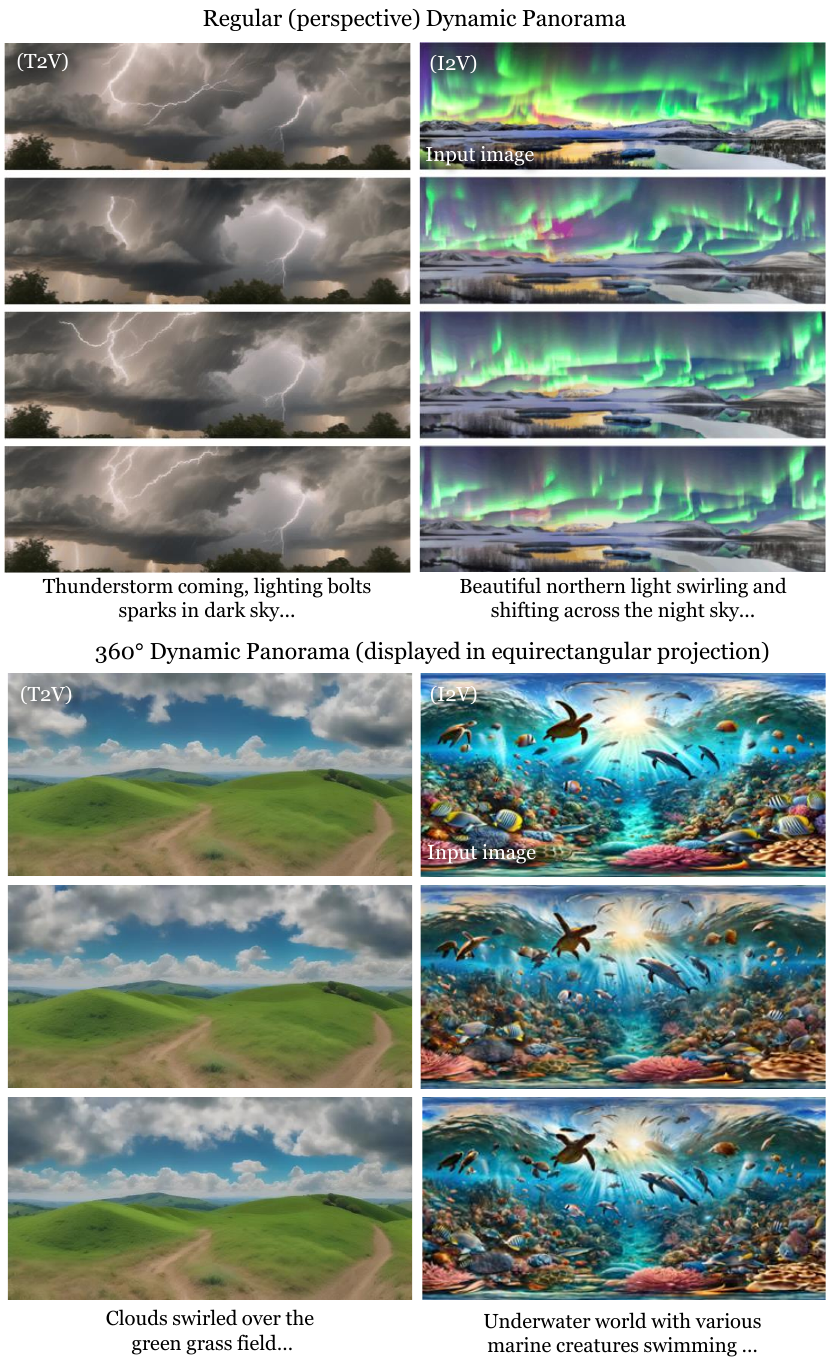} 
    \vspace{-0.45cm}
    \caption{We introduce \emph{\methodname}, a framework for generating dynamic panoramas conditioned on both images and text, or text alone. \emph{\methodname} enables the creation of regular panoramas with arbitary aspect ratio as well as 360° panoramic views, offering immersive visual experiences for AR/VR applications and displays of arbitrary aspect ratio and resolution.}
    \label{fig:title_fig}
    \vspace{-0.4cm}
\end{figure}

\end{abstract}

%% file: sec/1_intro.tex
\begin{figure*}[t]
    \centering
    \includegraphics[width=1\textwidth]{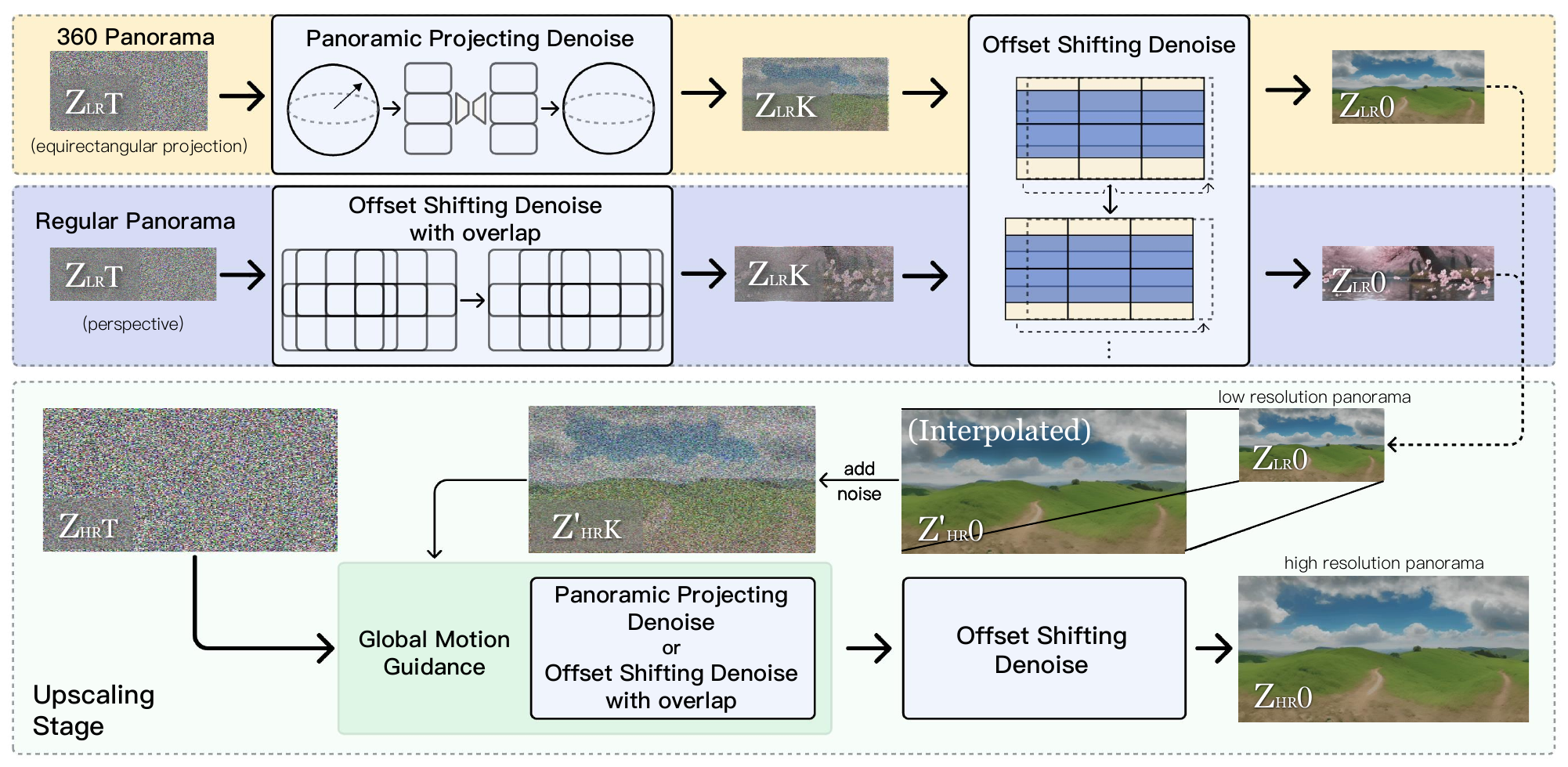} 
    \vspace{-0.4cm}
    \captionof{figure}{Our pipeline is divided into two stages: low-resolution stage establishes a coarse motion structure, 360-degree setting(the yellow block) involves Panoramic Projecting Denoise to initialize motion that fits to spherical panorama, while the regular perspective setting(the blue block) utilizes Offset Shifting with overlap for the early denoise steps, then the remaining denoise steps are completed by our Offset Shifting Denoise.  The up-scaling stage(the green block) utilizes more shift windows to produce a refined, high-resolution panorama with Global Motion Guidance from the low-resoltuion video.}
    \label{fig:pipeline}
    \vspace{-0.3cm}
\end{figure*}

\section{Introduction}

\label{sec:intro}
The increasing demand for immersive AR/VR applications has heightened the need for high-quality panoramic scene synthesis, essential for industries like digital advertising, wearable displays, and related tasks, where content often requires wide or portrait formats. However, achieving scalable panoramic scene synthesis poses unique challenges. A successful approach must enable spatially scalable generation while preserving motion coherence across panoramic scenes of any size, ensuring a seamless and immersive experience across various scenes in a panoramic view.

Recent methods for image generation face two key challenges: generating high-resolution or wide aspect ratio images, and maintaining motion consistency and memory efficiency in dynamic scene generation, such as video synthesis. Extending image generation to higher resolutions or wider aspect ratios is computationally intensive, requiring significant memory and large-scale training datasets.

Models like ~\cite{podell2023sdxl} support larger range of aspect ratios but still faces scalability issues, especially when it comes to ultra wide aspect ratio and higher resolution, which limits this methods in consistency, memory usage and inference speed. 
Other methods, such as those that stitch patches from pre-trained diffusion models, work well for generating panoramic or landscape images with repetitive patterns~\cite{lee2023syncdiffusion, bar2023multidiffusion}, but this approach is less effective for more complex or varied scenes. Additionally, although there have been advances in spatially scalable diffusion models for static image generation, these models are often limited to square images~\cite{du2024demofusion, graikos2024learned, guo2024make, zhang2023hidiffusion}, restricting their ability to handle broader aspect ratios. In contrast, the challenge of dynamic scene generation requires not only spatial coherence across frames, but also global motion consistency, making it even more computationally demanding. Moreover, video generation models must be designed with memory efficiency in mind, as large-scale dynamic scene synthesis often strains memory, limiting real-time inference capabilities. While recent works have explored training-free approaches for expanding diffusion models to new domains~\cite{voynov2023anylens, jin2023training, wang2023magicscroll}, the scalability of these models to high-resolution video generation remains largely underexplored, requiring solutions that balance motion consistency and memory consumption effectively.

Generating dynamic scenes in a 360° panoramic field of view (FoV) introduces unique challenges, including: (1) the broader content distribution required for equirectangular projections (ERPs) over 360° × 180° FoV; (2) curved motion patterns in ERPs versus straight-line motion in standard videos; and (3) continuity requirements at the left and right ERP boundaries, which represent the same meridian. 360DVD~\cite{wang2024360dvd} addresses these challenges by fine-tuning a text-conditioned video diffusion model on panoramic data in equirectangular space, but it suffers from low resolution and interpolation artifacts due to operating in the latent space, causing blurriness and divergence from the original motion space. Other methods, like 4K4DGen~\cite{li20244k4dgen} and Vividdream~\cite{lee2024vividdream}, attempt to animate scenes in overlapping regions, but their fixed-window denoising limits motion range and cross-scene consistency. Specifically, 4K4DGen faces constraints in motion range, relies solely on image-to-video transformations, and requires optimization procedures that reduce efficiency. Our approach overcomes these limitations by introducing a tuning-free denoising method that ensures spatial and temporal coherence for long-duration, loopable, and seamless panoramic video generation, achieving high-quality continuous motion with improved efficiency and visual fidelity.

We propose \textbf{\emph{\methodname}}, a unified, tuning-free framework for scalable panoramic dynamic scene synthesis with seamless motion. Our method ensures spatial and motion coherence throughout video generation by utilizing a shifting window that distributes noise uniformly across regions, achieving spatial scalability—whether overlapping or not—while maintaining consistent motion from a latent noise space. In contrast to the state-of-the-art 360DVD, \emph{\methodname} synthesizes higher-quality images by processing data into a pre-projected space before mapping it to the final equirectangular projection, thereby enhancing output fidelity. We introduce the Offset Shifting Denoiser (OSD), which synchronously denoises panoramic dynamic scenes using a well designed shifting Window mechanism, ensuring smooth transitions and spatial coherence while preserving motion fidelity as well as seamless transitions, and can also be adapted to generate 360 degree panorama by our Panoramic Projecting technique. To handle varying resolutions and aspect ratios, we employ Global Motion Guidance (GMG) and an upsampling strategy, ensuring local detail and global motion continuity in high-resolution scene generation. Our hierarchical approach maintains overall structure while delivering fine-grained local details, achieving seamless motion and scene-level consistency. In addition to spatial dimensions, we address the often-overlooked temporal dynamics, enabling the generation of long-duration as well as loopable dynamic scenes with continuous motion. We extend our OSD technique to the temporal domain, overcoming GPU memory constraints and enabling the synthesis of longer, temporally consistent dynamic scenes. Our contributions can be summarized as follows:
\begin{itemize} 
    \item We propose a unified framework for scalable panoramic dynamic scene synthesis, ensuring motion coherence across various resolutions, aspect ratios, and 360° FoV settings without requiring fine-tuning.
    \item We introduce the Offset Shifting Denoiser, which efficiently denoises the entire panoramic video with overall coherence, ensuring seamless boundary transitions and scene continuity across arbitrary aspect ratios, along with Global Motion Guidance, which enhances motion consistency at higher resolutions.
    \item We introduce the Panoramic Projection Denoiser to enable 360° FoV panorama generation and extend it to the temporal dimension, allowing for the generation of longer-duration or loopable dynamic videos. This method overcomes GPU memory limitations while ensuring temporal consistency across long-duration panoramic video sequences.
    \item Extensive experiments demonstrate that \emph{\methodname} outperforms existing methods in visual quality and motion consistency, generating continuous, longer duration an loopable dynamic scenes suitable for immersive applications.
\end{itemize}

%% file: sec/2_relatedwork.tex
\begin{figure*}[!htp]
    \centering
    \includegraphics[width=1\linewidth]{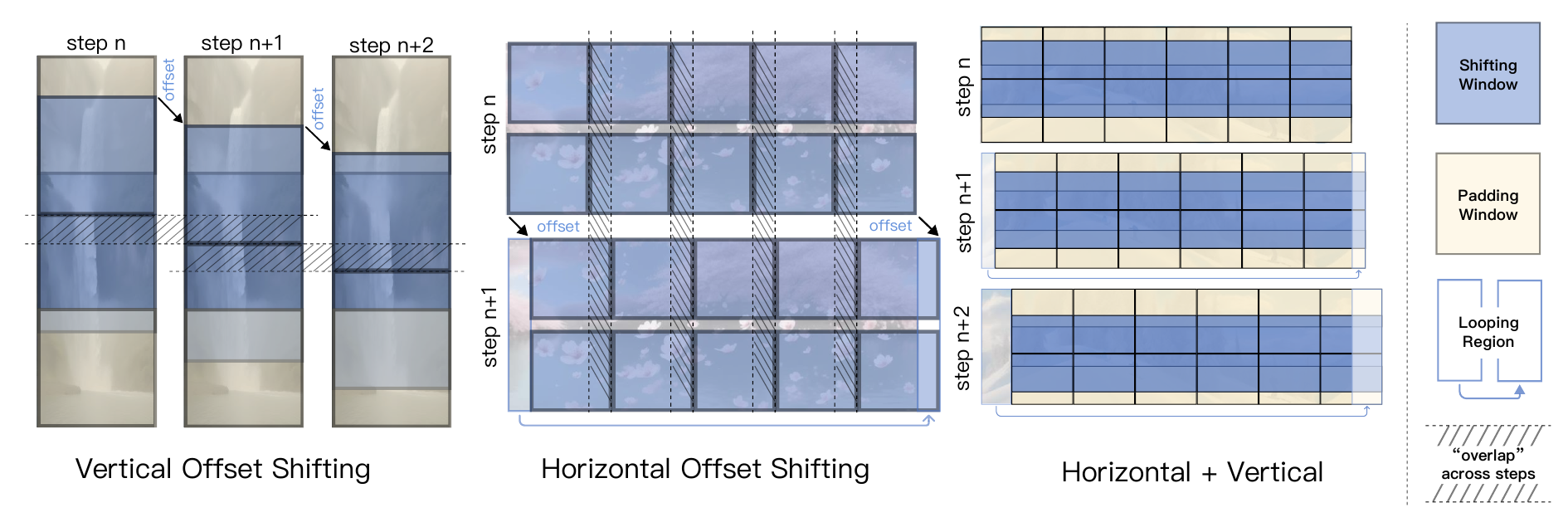}
    \vspace{-0.5cm}
    \caption{ 
    The purposed Offset Shifting Window mechanism, which involves shifting denoising windows both vertically and horizontally between denoise steps to denoise the whole panorama video latent with arbitrary aspect ratio an resolution. 
    The denosing windows are shifted vertically and horizontally every step, creating "overlap" regions between steps which mitigate the artifacts and synchronize the whole denoising process across the panorama. This results in seamless and consistent panoramic video generation with high resolution and aspect ratio. 
    }
    \label{fig:rec-pano-shift}
\end{figure*}

\section{Related Works}
\subsection{Spatial Scaling of Diffusion Models}
Diffusion models have achieved remarkable success in generating high-quality images, with recent efforts focusing on scaling to diverse resolutions and aspect ratios. Existing approaches often rely on retraining or fine-tuning large models~\cite{zheng2024any,xie2023difffit}, incurring significant computational costs and requiring extensive datasets. Other methods generate panoramic images by stitching patches from pretrained models, which work well for repetitive patterns like landscapes~\cite{lee2023syncdiffusion,bar2023multidiffusion,frolov2024spotdiffusion, quattrini2024merging}. Additionally, techniques such as AnyLens~\cite{voynov2023anylens}, MagicScroll~\cite{wang2023magicscroll}, and AutoDiffusion~\cite{li2023autodiffusion} extend pretrained models across various domains without retraining. However, most of these methods are limited to square aspect ratios or fixed resolutions~\cite{du2024demofusion,graikos2024learned,guo2024make,zhang2023hidiffusion} and cannot directly generate panoramic scenes with large aspect ratio. 

\subsection{Panoramic Scene Synthesis.}
Recent advancements in scene-level generation focus on synthesizing large-scale 3D scenes from text, as seen in works like LucidDreamer, GALA3D, and Wonderworld~\cite{chung2023luciddreamer,zhou2024gala3d,yu2024wonderworld}. These methods primarily generate static 3D scenes using Gaussian representations, limiting their capacity for dynamic content or flexible perspectives. Research on immersive generation has shifted toward 360° panoramas. OmniDreamer~\cite{akimoto2022diverse} employs cyclic inference for 360° image synthesis, while ImmenseGAN~\cite{dastjerdi2022guided} leverages fine-tuning for better control. Diffusion-based methods, such as DiffCollage~\cite{zhang2023diffcollage} and PanoDiff~\cite{wang2023360}, have shown promise for static panoramas, yet fail to address dynamic video generation. While DreamScene360~\cite{li2024dreamscene} integrates Gaussian splatting for text-to-panorama synthesis, its reliance on static priors restricts dynamic scene applications.

\subsection{Dynamic Scene Generation.}
Dynamic scene generation for panoramic videos introduces challenges in maintaining motion coherence, temporal consistency, and visual quality over extended durations. 360DVD~\cite{wang2024360dvd} adapts video diffusion models for panoramic data but faces limitations in generalization, interpolation accuracy, and style diversity when combining multiple LoRA models. Similarly, methods like 4K4DGen~\cite{li20244k4dgen} and Vividdream~\cite{lee2024vividdream} use overlapping regions for scene animation but suffer from noisy artifacts, fixed-window denoising, and restricted motion ranges. Moreover, 4K4DGen depends on optimization process to maintain coherence and is limited to image-to-video transformations. In contrast, our method overcomes these limitations with a shifting-based denoising framework that ensures spatial and temporal coherence. Our approach enables the generation of seamless, long-duration, and loopable panoramic videos with high-quality dynamic motion.

%% file: sec/3_method.tex
\section{Method}

We present \emph{\methodname}, a scalable and tuning-free framework for panoramic dynamic scene synthesis that achieves seamless spatial and temporal coherence in panoramic sythesis, as illustrated in Figure~\ref{fig:pipeline}. To extend video diffusion from fixed resolutions to expansive panoramas, our approach involves an Offset Shifting Denoising mechanism. To further enhance structural awareness and ensure global coherence in large-scale motion, we introduce Global Motion Guidance, a structured motion prior tailored for panoramic video generation. Additionally, our framework provides methods for continuous, loopable motion, enabling seamless transitions and frame-to-frame consistency for extended, loopable video sequences.

\subsection{Offset Shifting Denoising}

To generate a $W_p \times H_p$ panorama $Z$ with a regular diffusion model $\theta$ trained in resolution $W_\theta \times H_\theta ( W_\theta < W_p, H_\theta < H_p )$ without extra finetuning, $Z$ should be divided into $n_W \times n_H$ windows to fit in the applicable size and denoised. Existing methods in panorama image synthesis like MultiDiffusion~\cite{bar2023multidiffusion} and SyncDiffusion~\cite{lee2023syncdiffusion} utilize more windows $ (n_H > H_p / H_\theta, n_W > W_p / W_\theta) $ that overlap with each others to mitigate the seam and distortion caused by divided denoising windows, resulting in much higher computation overhead, while ~\cite{frolov2024spotdiffusion} randomly shift the windows to reduce the overlap needed, those methods mainly focus on regular perspective panorama image. However, in high resolution 360 degree panoramic video synthesis require extending both vertically and horizontally, contents also differs between each windows, hindering one from directly applying those methods. To tackle this challenge, we introduce Offset Shifting Denoising (OSD) that shift denoising windows by an offset, both vertically and horizontally, across the entire panorama in each steps as illustrated in Figure~\ref{fig:rec-pano-shift}. This offset creates "overlap" regions among windows between denoise steps, synchronizing content and motions across windows, and the trend of discontinuity at the edge of windows in one certain denoise step are also mitigated at the next step because the edges are also shifted along with the window. 

Specifically, in the vertical direction, the windows with in the upper and lower boundaries are shifted vertically by an offset every steps, and the region that is not covered by those \textbf{shifting windows} are handled by the \textbf{padding windows} at the top and bottom of the panorama. Horizontally, windows are also shifted by offset while the whole latent would be regarded as a ring, that is, the left and right boundaries are 'connected' so windows can cross through those boundary, which is implemented by filling the 'out of boundary' regions by the corresponding region from the other side. The Offset Shifting Denoising process can be formulated as:
\begin{equation}
\label{eq:OSD}
Z_t = Con|_{1:n_W, 1:n_H}( \Phi_\theta(t,c,Split(Z_{t-1},i,j,t,n_W,n_H)))
\end{equation}
where $Split(X,i,j,t,n_W,n_H)$ indicates dividing the panorama video latent $X$ into $n_W \times n_H$ windows with offset at time $t$ and select the one in $i-th$ column, $j-th$ row; 
$Con(\cdot)$ means collecting latent from divided windows and concat them to form the entire panorama latent; 
$\Phi_\theta(t,c,x)$ represents the diffusion process, for T2V, $c = \{c_{text}\}$; and for I2V, $c = \{c_{text}, Split(c_{img},i,j,t)\}$. 

OSD not only allows seamless transition across the whole generated video in arbitrarily aspect ratio but also enables motion and content continuity between the left and right boundaries, as shown in Figure~\ref{fig:seemless_sample}, producing immersive panoramic videos.


\subsection{Global Motion Guidance}

Shifting denoising along horizontal and vertical directions enables scalable panorama generation and seamless motion. However, complex motion patterns, such as cascading waterfalls, require coordination across the entire panoramic field.
At early denoising steps, the overall layout are constructed ~\cite{patashnik2023localizing}, 
while the intrinsic synchronization offered by OSD has not accumulate enough influence in this period, different part of the panorama may tend to separate motion pattern, resulting in less consist overall motion, especially in high resolution panoramic scene generation.
Besides applying more windows to create explicit spatial overlap in the first $K$ steps, we introduce the Global Motion Guidance (GMG), which decompose the generation process into global layout and local content stages hierarchically: the first stage synthesis a video in lower resolution, capturing high-level motion structures. Then in the upscaling stage those low-resolution result are upsampled using interpolation algorithm \( {inter}(\cdot) \) like bicubic and re-noised \( {noise}(\cdot) \), serving as an initialization that guide content layout and motion in the following high-resolution generation, which refines local details while preserve consistency. This process is formulated as: 
\begin{equation}
{Z_{HR^0}} =  {\Phi_\theta}^{OSD} ( {noise}( {inter} ( {\Phi_\theta}^{OSD}(Z_{LR^T}) )) ) 
\end{equation}

Integrated with Offset Shifting Denoise, Global Motion Guidance preserves both broad motion structures and intricate details, yielding cohesive and richly detailed high resolution panoramic scenes.

\begin{figure}[!htp]
    \centering
    \includegraphics[width=1\linewidth]{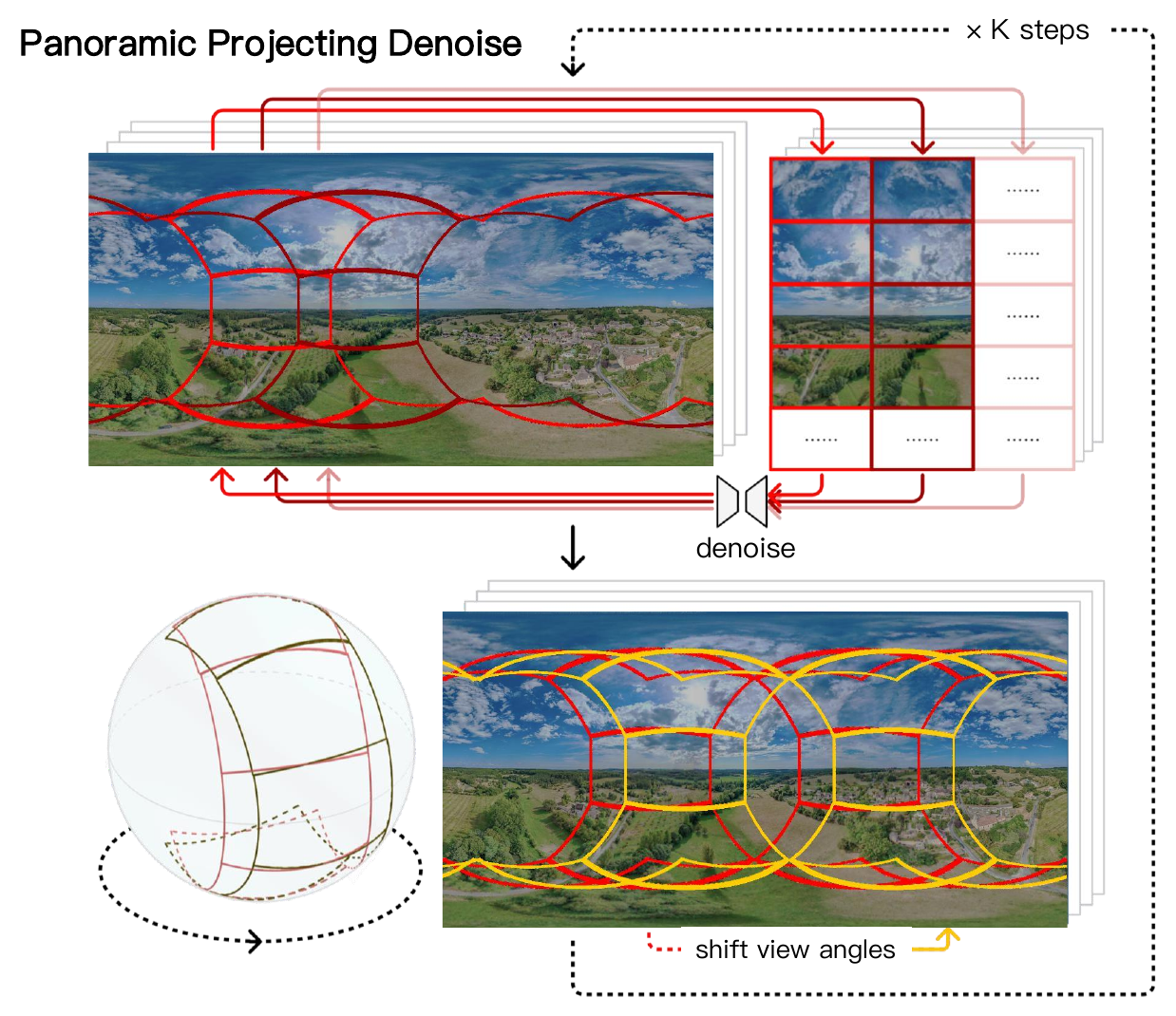}
    \vspace{-0.5cm}
    \caption{The purposed Panoramic Projecting Denoise, where spherical panorama videos latents (represented as equirectangular projections) are projected into perspective view port windows and denoised, followed by re-projection back to the equirectangular panorama, as shown in the upper part of the figure. Those view port windows are also shifted with an offset applied in their view angles at each steps, as shown in the lower part of the figure. For legibility, only a proportion of view port regions are shown in the figure.}
    \label{fig:pano-proj}
    \vspace{-0.5cm}
\end{figure}

\subsection{360° FoV Panorama Generation}

360° FoV panoramas are usually represented by its equirectangular projection (ERP), which project the spherical panorama into an \( H \times W \times C \) matrix, with \( W / H = 2\). Common diffusion models trained by regular (perspective) video datasets face challenge in generating 360° FoV panorama videos due to the deformation in equirectangular projection. Besides training adapters with extra 360° FoV datasets ~\cite{wang2024360dvd}, projecting portion of the equirectangular panorama back into mulitple perspective view ports allows using regular diffusion models to denoise. ~\cite{li20244k4dgen} relys on an optimization procedure to synchronize the diffusion process among each view ports, while it is constrained in Image to Video Generation and suffer from insufficient motion. We instead intergrate our offset shifting denoising mechanism with equirectangular projecting, ensuring high-quality 360° FoV panorama generation with minimal distortion.

\subsubsection{Spherical Projection.}
Given a point $(x, y, z)$ on the unit sphere, the corresponding longitude $\alpha$ and latitude $\beta$ can be computed by:
\begin{equation}
    \alpha = \arctan2(y, x), \; \beta = \arcsin(z).
\end{equation}

The equirectangular projection maps these spherical coordinates $(\alpha, \beta)$ to a rectangular plane $(u, v)$  with $u = \alpha,  v = \beta$,
where the rectangular plane has a width of $2\pi$ and a height of $\pi$, maintaining a $1:2$ aspect ratio.
For the inverse mapping, given a point $(u, v)$ on the rectangular plane, the corresponding 3D coordinates $(x, y, z)$ on the unit sphere can be computed by:
\begin{equation}
    x = \cos(\beta) \cos(\alpha), \; y = \cos(\beta) \sin(\alpha), \; z = \sin(\beta).
\end{equation}

\subsubsection{Offset Shifting in Panoramic Space.}  

Perspective view ports at certain view angle $a$ and field of view $f$ can be obtained by the projecting function $ Proj(\cdot)$ from the equirectangular projection $Z^p$ of a 360 degree panorama, dividing the panorama into $ n_\alpha \times n_\beta $ view port windows that can be denoised using normal diffusion model $ \theta $. Similar to the Offset Shifting in perspective panorama, we can also shift the view port windows at each steps with an offset applied in view angle $a$, as illustrated in Figure~\ref{fig:pano-proj}, which can be formulated as: 
\begin{equation}
Z_t = Con_{Proj}|_{1:n_\alpha, 1:n_\beta}(\Phi_\theta(t,c,Proj(Z_{t-1}, f, a_{i,j,t})))
\end{equation}
where $Con_{Proj}(\cdot)$ means collecting latent from divided view port windows and project into the 360 degree panorama.

In cases that windows have overlap with each others, which is often the case in high polar angle regions, parts of the latent space corresponding to subsequent windows may have already been denoised by previous windows, creating inconsistent noise levels. To address this, we maintains a mask $M_d$ at each timestep $t$ to track denoised regions. Before processing each window, noise is rebalanced as:
\begin{equation}
Z'_t[x, y] = 
    \begin{cases} 
    \sqrt{\alpha_t}Z_t[x, y] + \sqrt{1-\alpha_t}\epsilon_t & \text{if } M_d(x,y) = 1 \\
    Z_t[x, y] & \text{if } M_d(x,y) = 0
    \end{cases}    
\end{equation}
ensuring consistent noise levels across all windows before passing into diffusion model. 
\begin{figure}[htp]
    \centering
    \includegraphics[width=1\linewidth]{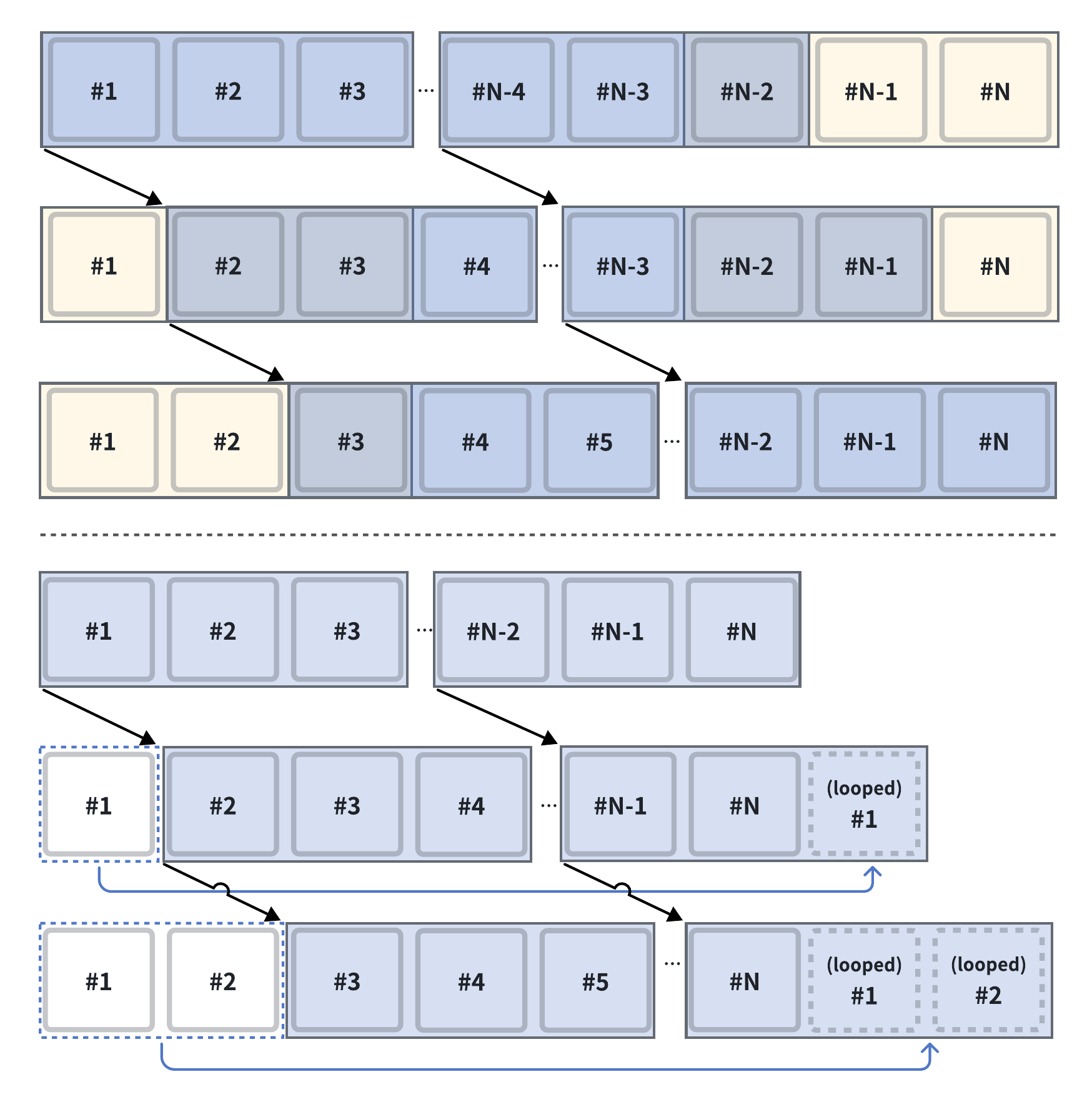}
    \vspace{-0.3cm}
    \caption{The Offset Shifting Denoising mechanism extended to temporal dimension. The upper part shows how the frame clip windows are shifted with an offset along the temporal dimension, with padding windows at the start and end of the video sequence. The lower part shows the loopable offset shifting denoising, with looping frames at the start and end of the frames sequence.}
    \label{fig:temporal_offset_shifting}
    \vspace{-0.5cm}
\end{figure}

\subsection{Long and Loopable Scene Video Generation}

\begin{figure}[htp]
    \vspace{-0.4cm}
    \centering
    \includegraphics[width=1\linewidth]{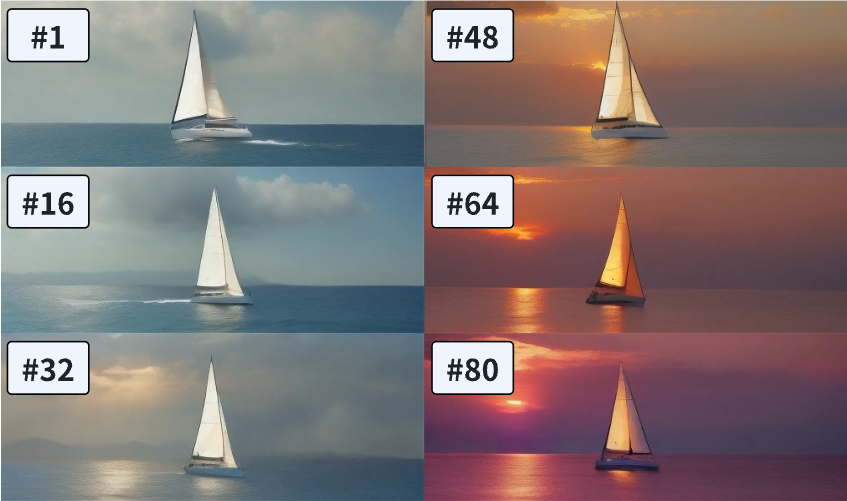}
    \vspace{-0.4cm}
    \caption{Example frames from generated video at the first, 16th, 32th, 48th, 64th and 80th frames generate by a diffusion model that is capable for 16 frames originally. Despite the increasing video length, the visual quality of the panorama remains consistent, demonstrating the effectiveness of our method in generating long videos. }
    \label{fig:long_video}
    \vspace{-0.3cm}
\end{figure}

Most existing dynamic scene synthesis approaches like ~\cite{wang2024360dvd} and ~\cite{li20244k4dgen} are constrained in limited duration video sequences, as most video diffusion models produce only short clips (typically 16 frames, 2s durantion at 8 FPS). To address this, we extended Offset Shifting Denoising from spatial to temporal dimension and introduced a temporal shifting strategy that enables continuous motion and the generation of videos with much longer durationby regular video diffusion models $ \theta $ that have limited frames length capacity $ F_\theta $. In detail, the latent of long duration video $ Z \in \mathbb{R}^{F' \times W \times H}$ is divided along temporal dimention $ F' $  into $ n_F$ shorter frame clip windows with duration $ F_\theta $ each, as shown in Figure~\ref{fig:temporal_offset_shifting}. which can be formulated by:
\begin{equation}
\label{eq:temp_OSD}
Z_t = Con_F|_{1:n_F}(\Phi_\theta(t,c,Split_F(Z_{t-1},i,t,n_F)))
\end{equation}

where $Split_F(X,i,j,t,n_F)$ indicates dividing the panorama video latent $X$ into $n_F$ frame clip windows with offset at time $t$ and select the $i-th$ one; $Con_F(\cdot)$ means collecting latent from divided windows and concat them to form the complete latent. 
The shifting offset in temporal dimensions also create 'cross step overlap' between windows analogous to spatial OSD, enabling generation of videos with much longer duration by regular video diffusion models that have limited frame length, while perserving continuous motion and smooth transitions between frames, as shown in Figure~\ref{fig:long_video}. 

Moreover, by applying similar looping mechanism in our spatiall OSD process to the temporal dimension we can achieve seamless looping video generation. Specifically, instead of the padding windows at the start and the end of the frame sequence, the first and last frame are regarded as connected and windows are allowed to cross the start and end boundary, treating the video frames as a ring, to create a loopable sequence with smooth, continuous motion.

%% file: sec/4_exp.tex
\begin{table*}[htp]
    \centering
    \small
    \renewcommand{\arraystretch}{0.99} 
    \setlength{\tabcolsep}{3pt} 
    \begin{tabular}{lcccccccc}
    \hline
     & Source & \thead{Tuning-Free} & \thead{Arbitrary \\ Size} & \thead{360° \\ Field-of-View} & \thead{Text Only \\ Condition} & \thead{Image \\ Condition} & \thead{Unlimited \\ Video Length} & \thead{Loopable \\ Generation} \\
    \hline
    360DVD~\cite{wang2024360dvd} & CVPR24 & $\times$ & $\checkmark$ & $\checkmark$ & $\checkmark$ & $\times$ & $\times$ & $\times$  \\
    4K4DGen~\cite{li20244k4dgen} & Arxiv24 & $\times$ & $\times$ &  $\checkmark$ & $\times$ & $\checkmark$ & $\times$ & $\times$ \\
    Scalecrafter~\cite{he2023scalecrafter} & ICLR24 & $\checkmark$ & $\checkmark$ & $\times$ & $\checkmark$ & $\times$ & $\times$ & $\times$ \\
    VividDream~\cite{lee2024vividdream} & Arxiv24 & $\times$ & $\checkmark$ & $\times$ & $\times$ & $\checkmark$ & $\times$ & $\times$ \\
    DynamicScaler & - &$\checkmark$ & $\checkmark$ & $\checkmark$ & $\checkmark$ &  $\checkmark$ & $\checkmark$ &  $\checkmark$ \\
    \hline
    \hline
 &  \thead{CLIP-Score}$\uparrow$ & \thead{Image \\ Quality}$\uparrow$& \thead{Dynamic \\ Degree}$\uparrow$ & \thead{Motion \\ Smoothness}$\uparrow$ & \thead{Temporal \\ Flickering}$\uparrow$ 
    & \thead{Scene}$\uparrow$ & \thead{Q-Align(I)}$\uparrow$ & \thead{Q-Align(V)}$\uparrow$    \\
    \hline
    360DVD~\cite{wang2024360dvd}  & 0.293 & 0.436 & 0.412 & 0.917 & 0.964 & 0.417 & 0.485 & 0.532 \\
    DynamicScaler  & \textbf{0.302} & \textbf{0.583} & \textbf{0.783} & \textbf{0.963} & \textbf{0.982} & \textbf{0.499} & \textbf{0.632} & \textbf{0.613} \\
    \hline
    \end{tabular}
    \caption{    
    Quantitative comparison of dynamic scene generation methods, with best results highlighted in bold. The evaluation covers key factors such as resolution scalability, video length, and loopability, using metrics on image quality, dynamic range, motion smoothness, and temporal flickering, and user-centric Q-Align scores. DynamicScaler outperforms existing methods across all these metrics.}
    \label{tab:quantitative_comparison}
\end{table*}

\section{Experiments}

\subsection{Qualitative Results}

\begin{table}[htp]
\centering
\small
\resizebox{0.5\textwidth}{!}{ 
\renewcommand{\arraystretch}{1.0} 
\setlength{\tabcolsep}{1pt} 
\begin{tabular}{lcccccc}
\hline
  \multicolumn{1}{c}{}  & \multicolumn{2}{c}{Video Criteria} & \multicolumn{3}{c}{Panorama Criteria} \\
    \cline{2-3} \cline{4-6}
    Methods & \thead{Graphics \\ Quality}$\uparrow$ & \thead{Frame \\ Consistency}$\uparrow$ & \thead{End \\ Continuity}$\uparrow$ & \thead{Motion \\ Pattern}$\uparrow$ & \thead{Scene \\ Richness}$\uparrow$ \\ 
    \hline
    \multicolumn{6}{l}{\textbf{Same Case Comparison}} \\
    360DVD~\cite{wang2024360dvd} & 3.3 & 3.5 & 3.6 & 3.7 & 3.5 \\
    Ours & \textbf{4.6} & \textbf{4.7} & \textbf{4.8} & \textbf{4.5} & \textbf{4.6} \\
    \hline
    \multicolumn{6}{l}{\textbf{Random Case Comparison}} \\
    360DVD~\cite{wang2024360dvd} & 3.3 & 3.4 & 3.6 & 3.9 & 3.4 \\
    4K4DGen~\cite{li20244k4dgen} & \textbf{4.5} & 3.6 & 4.3 & 3.6 & 4.3 \\
    Scalecrafter~\cite{he2023scalecrafter} & 3.5 & 3.7 & 1.9 & 4.4 & 3.6 \\
    VividDream~\cite{lee2024vividdream} & 3.6 & 3.7 & 3.8 & 3.6 & 4.1 \\
    Ours & 4.3 & \textbf{3.9} & \textbf{4.5} & \textbf{4.5} & \textbf{4.4} \\
    \hline
\end{tabular}
} 
\caption{User preference study results. \textbf{Same Case Comparison} refers to using the same case for comparison, while \textbf{Random Case Comparison} refers to using different cases due to the unavailability of some methods. Ratings range from 1 (lowest) to 5 (highest).}
\label{tab:user_preference_studies}
\vspace{-0.6cm}
\end{table}

\begin{table}[htp]
\resizebox{0.5\textwidth}{!}{
    \centering
    \setlength{\tabcolsep}{0.2pt} 
    \begin{tabular}{ccccccc}
        \hline
        & \thead{Image \\ Quality}$\uparrow$ & \thead{Dynamic \\ Degree}$\uparrow$ & \thead{Motion \\ Smoothness}$\uparrow$ & \thead{Temporal \\ Flickering}$\uparrow$ & \thead{Q-Align(V)}$\uparrow$ \\
        \hline
        {Direct Inference} & OOM & OOM & OOM & OOM & OOM \\
        {w/o OSD} & 0.564 & 0.749 & 0.948 & 0.905 & 0.595 \\
        {w/o GMG} & 0.571 & 0.765 & 0.961 & 0.946 & 0.598 \\
        {Full Method} & \textbf{0.587} & \textbf{0.778} & \textbf{0.967} & \textbf{0.985} & \textbf{0.616} \\
        \hline
    \end{tabular}
    }
    \caption{
    Performance comparison of different configurations regarding image quality, dynamic degree, motion smoothness, and temporal flickering. OOM stands for Out-of-Memory. 
    }
    \label{tab:ablation_study}
    \vspace{-0.4cm}
\end{table}

We evaluate the performance of \methodname against state-of-the-art models. We detail the implementation, experimental settings, comparison methods, and user studies conducted to assess the efficacy of our approach. The proposed \methodname is built on top of existing text-to-video generation model from VideoCrafter2~\cite{chen2024videocrafter2} and image-to-video generation model from VideoCrafter1~\cite{chen2023videocrafter1}.  

Compared to previous methods, \methodname supports a wider range of settings and can generate videos of infinite length, while other methods are limited to generating only finite-length videos. Our approach generates dynamic panoramas at various resolutions, including both standard and 360° Field-of-View (FoV) panoramas, with support for both text- and image-conditioned inputs, as shown in Table~\ref{tab:quantitative_comparison}. Currently, among the available concurrent works, only 360DVD is capable of generating dynamic panoramas, so we compare our method directly with it. We assess the visual quality by randomly selecting views from 100 generated video cases using random camera positions. Key factors such as image quality, dynamic range, motion smoothness, temporal flickering, and scene richness are evaluated according to the protocols of VBench~\cite{huang2024vbench} and CogVideoX~\cite{yang2024cogvideox}. Furthermore, we incorporate an LLM-based visual evaluator, Q-Align~\cite{wu2023q}, to score both image and video quality. As shown in Table~\ref{tab:quantitative_comparison}, \emph{\methodname} outperforms existing methods, demonstrating superior video quality in handling complex dynamic scenes.

\begin{figure}[!h]
    \centering
    \includegraphics[width=0.9\linewidth]{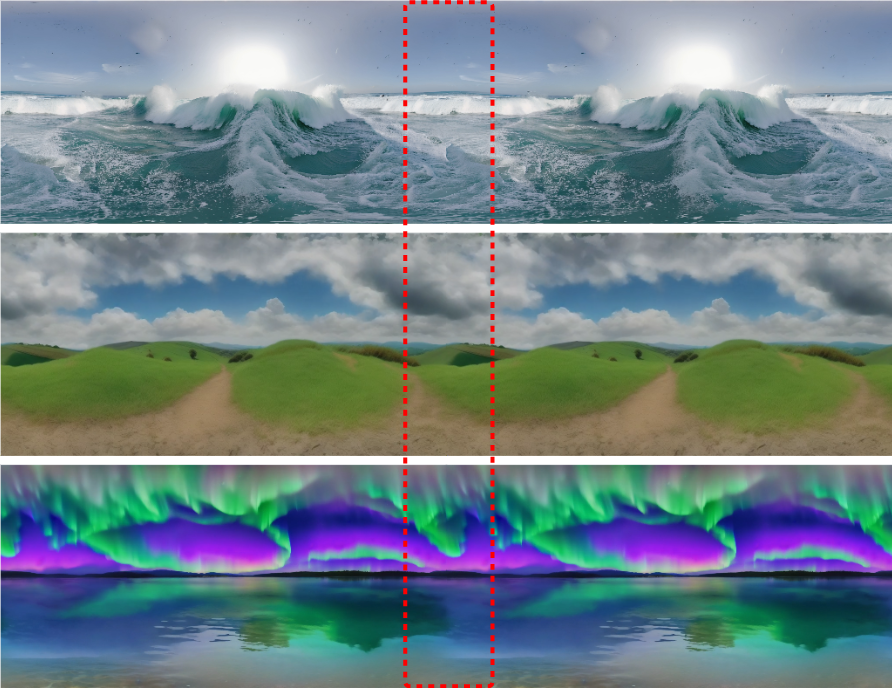}
    \caption{To demonstrates the seamlessness of generated panorama videos, we horizontally concatenat the video frames, showcasing continuity across the left and right boundaries.}
    \label{fig:seemless_sample}    
    \vspace{-0.5cm}

\end{figure}

\begin{figure*}[htp]
    \centering
    \includegraphics[width=0.99\linewidth]{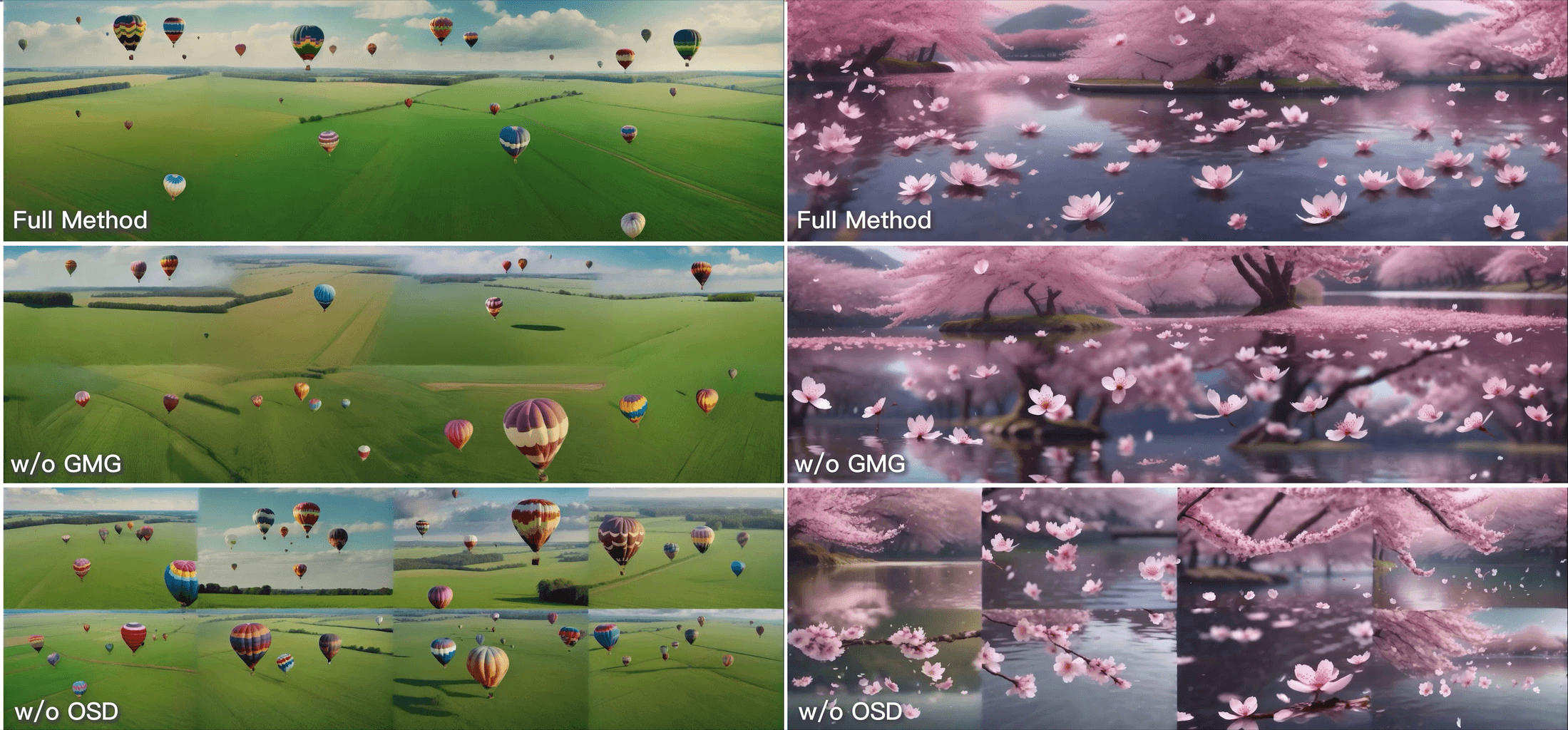}
    \vspace{-0.1cm}
    \caption{Qualitative visualization of ablation study in regular perspective setting. Refer to our project page for results in video format.}
    \label{fig:scalability-results}
    \vspace{-0.4cm}
\end{figure*}

\subsection{Quantitative Results}

As demonstrated in Figure~\ref{fig:title_fig}, our approach generates panoramas of arbitrary sizes, supporting rectangular and 360° Field of View (FoV) configurations, offering unparalleled flexibility for diverse applications. As shown in Figure~\ref{fig:seemless_sample}, the region near horizontal boundaries of generated frames align seamlessly, ensuring flawless 360° view—an essential feature for immersive environments. This highlights the robustness of our shift denoise technique in both non-360° and 360° settings. Furthermore, our method facilitates the creation of scene-level, long duration, and loopable videos through temporal shifting, without requiring additional training or compromising visual quality. 
\vspace{-0.1cm}

\subsection{User Studies}
We conduct user studies to evaluate videos based on five criteria: graphics quality, frame consistency, left-right continuity, content distribution, and motion patterns. 
20 Participants are asked to select the video with the highest quality.
Table~\ref{tab:user_preference_studies} shows that \methodname performs favorably in all criteria. 
We use two comparison types: same-case comparison (using the same scene for direct comparisons) and random-case comparison (using different scenes due to limited publicly available samples for comparison). \emph{\methodname} receives the highest ratings in most criteria as shown in Table ~\ref{tab:user_preference_studies}, especially for continuity, content distribution, and motion patterns, demonstrating its superior ability to generate seamless, dynamic panorama.

\subsection{Ablation Studies}

We conducted ablation studies to evaluate the impact of each core component in \methodname under four different configurations: \textbf{Direct Inference}, where the video is generated directly at the target resolution without enhancement techniques, but results in out-of-memory (OOM) due to high VRAM demands; \textbf{Without OSD (Offset Shifting Denoiser)}, which excludes the offset spatial denoising mechanism that mitigates multi-scale artifacts; \textbf{Without GMG (Global Motion Guidance)}, where global motion guidance is omitted, reducing frame-to-frame motion continuity; and the \textbf{Full Method}, which integrates all components. As shown in Table~\ref{tab:ablation_study}, the Full Method consistently outperforms all other configurations across all evaluation metrics, underscoring the significance of each component. Both OSD and GMG play critical roles in improving image quality, dynamic range, motion smoothness, and temporal coherence, while also minimizing temporal flickering. We also provide quantitative results of ablation study in Figure~\ref{fig:scalability-results} and our project page.

\section{Conclusions}
This paper presents \methodname for scalable, coherent panoramic dynamic scene synthesis. By introducing Offset Shifting Denoiser, our approach ensures efficient denoising and consistent boundary transitions. And Global Motion Guidance mechanism maintains local detail and global motion continuity, delivering superior content quality and motion smoothness. 

Overall, \methodname outperforms existing methods in scalability and performance, offering a practical, training-free solution for creating high-quality, immersive AR/VR dynamic content across various resolutions and aspect ratios.

\section*{Acknowledgments}
Supported by the Intelligence Advanced Research Projects Activity (IARPA) via Department of Interior/ Interior Business Center (DOI/IBC) contract number 140D0423C0074. The U.S. Government is authorized to reproduce and distribute reprints for Governmental purposes notwithstanding any copyright annotation thereon. Disclaimer: The views and conclusions contained herein are those of the authors and should not be interpreted as necessarily representing the official policies or endorsements, either expressed or implied, of IARPA, DOI/IBC, or the U.S. Government.